\PassOptionsToPackage{table}{xcolor} 
\documentclass[10pt,twocolumn,letterpaper]{article}

\usepackage{iccv}              

\usepackage{multirow}

\definecolor{iccvblue}{rgb}{0.21,0.49,0.74}
\usepackage[pagebackref,breaklinks,colorlinks,allcolors=iccvblue]{hyperref}

\usepackage{float}
\usepackage{multirow}
\usepackage{stfloats}
\usepackage{xcolor} 

\def\paperID{11} 
\def\confName{ICCV}
\def\confYear{2025}

\title{Stable-Drift: A Patient-Aware Latent Drift Replay Method for Stabilizing Representations in Continual Learning}

\author{Paraskevi-Antonia Theofilou$^{1}$ \and Anuhya Thota$^{2}$ \and Stefanos Kollias$^{1}$  \and  Mamatha Thota$^{3}$ \and  $^1$National Technical University of Athens, Greece 
\and \and $^{2}$London School of Economics and Political Science, UK \and $^{3}$University of Lincoln, UK \and
{\tt\small partheofilou@ails.ece.ntua.gr, a.thota1@lse.ac.uk, stefanos@cs.ntua.gr, mthota@lincoln.ac.uk}
}

\begin{document}
\maketitle
\begin{abstract}
When deep learning models are sequentially trained on new data, they tend to abruptly lose performance on previously learned tasks, a critical failure known as catastrophic forgetting. This challenge severely limits the deployment of AI in medical imaging, where models must continually adapt to data from new hospitals without compromising established diagnostic knowledge. To address this, we introduce a latent drift-guided replay method that identifies and replays samples with high representational instability. Specifically, our method quantifies this instability via "latent drift", the change in a sample's internal feature representation after naive domain adaptation. To ensure diversity and clinical relevance, we aggregate drift at the patient level; our memory buffer stores the per patient slices exhibiting the greatest multi-layer representation shift. Evaluated on a cross-hospital COVID-19 CT classification task using state-of-the-art CNN and Vision Transformer backbones, our method substantially reduces forgetting compared to naive fine‑tuning and random replay. This work highlights latent drift as a practical and interpretable replay signal for advancing robust continual learning in real-world medical settings.
\end{abstract}
\section{Introduction}
\label{sec:intro}

Deep learning has demonstrated remarkable performance in medical image analysis, providing automated solutions for critical tasks such as disease detection, diagnosis support, and patient stratification \cite{litjens2017survey, esteva2019guide}. However, the deployment of deep neural networks in real-world clinical workflows remains challenging due to the dynamic and heterogeneous nature of medical data. Hospitals and imaging centers often differ in acquisition protocols, scanner hardware, patient demographics, and disease prevalence, resulting in significant domain shifts that undermine the generalization ability of conventional models trained on static datasets \cite{chen2019domain, wang2020generalizing}.

To address this, continual learning (CL) has emerged as a promising paradigm that enables models to incrementally adapt to new data distributions while preserving previously learned knowledge \cite{thota2023lleda, delange2021continual}. Among CL strategies, replay-based methods, which store and revisit a small subset of past data, are highly effective. However, the efficacy of replay is critically dependent on the composition of the memory buffer, and many existing approaches rely on naive random sampling, which is often suboptimal and fails to store the most critical information needed to prevent forgetting. A major obstacle thus remains: catastrophic forgetting, where fine-tuning on new data leads to a severe degradation of performance on earlier domains \cite{kirkpatrick2017overcoming, mccloskey1989catastrophic}. This is particularly problematic in medical imaging, where retaining established diagnostic knowledge is essential for patient safety and model trustworthiness \cite{chen2021continual}.

In this paper, we propose a novel, latent-drift–guided replay framework to mitigate catastrophic forgetting. Our approach is founded on the principle that samples most susceptible to forgetting are those whose internal feature representations become most unstable during domain adaptation. We quantify this instability by calculating the \textit{latent drift}, the change in a sample's representation between a model trained on the source domain and one naively fine-tuned on the target. By identifying samples with the highest latent drift, we construct an intelligent replay buffer designed to preserve the most fragile knowledge. Crucially, our method aggregates drift scores at the patient level and across multiple model layers, ensuring that the buffer is not only informative but also diverse and clinically relevant.

We conduct extensive experiments on a real-world, cross-hospital COVID-19 CT dataset, evaluating our approach with both a state-of-the-art CNN (ResNet50, \cite{he2016deep}) and a Vision Transformer (Swin Transformer, \cite{liu2021swin}) backbone. Our results demonstrate that our latent drift-guided strategies significantly outperform naive fine-tuning and random replay baselines, establishing a new state-of-the-art for this task. In summary, our main contributions are:
\begin{itemize}
\item We introduce latent representation drift as a practical and interpretable signal for identifying samples at high risk of being forgotten during continual learning in medical imaging.

\item We propose a novel patient-aware, multi-layer buffer selection strategy that leverages drift signal to construct a compact and highly effective replay memory.

\item Our framework sets a new benchmark for cross-domain continual learning on a challenging COVID-19 dataset, offering a robust solution that balances knowledge retention and adaptation.

\end{itemize}

\section{Related work}
\label{sec:related}

\subsection{Continual Learning beyond Domain Adaptation}

In many real-world scenarios, models must adapt to shifts in data distribution, a challenge addressed by domain adaptation (DA), which focuses on transferring knowledge from a source domain to a different but related target domain \cite{thota2021contrastive, thota2020multi}. DA typically assumes access to both domains during training, continual learning (CL), or lifelong learning, extends this idea by requiring models to sequentially learn from new domains or tasks over time without forgetting previously acquired knowledge \cite{thota2023lleda, parisi2019continual}.
A central challenge in CL is catastrophic forgetting, where fine-tuning on new data overwrites or degrades earlier representations \cite{mccloskey1989catastrophic}.

To address this, researchers have explored strategies such as regularization-based methods \cite{kirkpatrick2017overcoming}, parameter isolation \cite{mallya2018packnet}, and replay-based approaches \cite{rebuffi2017icarl}. Replay-based methods store selected samples from prior tasks and mix them with new data during training. This simple yet effective strategy has shown success in domains such as computer vision \cite{lopez2017gradient}, natural language processing \cite{shin2017continual}, and medical imaging \cite{chaudhry2019efficient}. However, deciding which samples to store and how to manage the replay buffer remains an open problem: Random sampling is common but may overlook the most informative or diverse examples.

\subsection{Replay Buffer Management}
Recent studies have explored more intelligent buffer construction strategies to maximize diversity and informativeness. Gradient-based methods (e.g., GSS \cite{aljundi2019gradient}) 
prioritize samples with high influence on model updates, while uncertainty-based approaches use model confidence \cite{farquhar2018towards}. Other works such as MIR \cite{ aljundi2019online} retrieve samples with maximal gradient interference, ESMER \cite{ sarfraz2023error} favors low-loss “anchors” to counter abrupt drift, and LDC \cite{ gomez2024exemplar} learns to compensate drift via an auxiliary module. Latent-drift measures have also been used to track representation change \cite{ kornblith2019similarity}, but not for replay selection with patient-aware constraints.

Our method differs by computing multi-layer latent drift between two domain-specific model states, aggregating at the patient level, and enforcing class-balanced replay, yielding targeted retention under cross-hospital domain shift without enlarging the buffer.

\subsection{Continual Learning in Medical Imaging}
Medical imaging poses unique challenges for CL due to domain shifts caused by varying scanner hardware, acquisition protocols, and patient populations \cite{chen2021continual}. Moreover, the high stakes in clinical decision-making necessitate robust retention of prior knowledge. Several works have explored CL in medical image analysis, focusing on segmentation \cite{dou2019domain}, classification \cite{shmelkov2017incremental}, and detection \cite{li2019learning}.

Replay-based techniques have been applied to medical imaging to address domain adaptation and data privacy constraints \cite{islam2021continual}. However, most existing approaches rely on random or heuristics-based sample selection for the replay buffer. Our method differs by employing latent drift-informed buffer management tailored to multi-hospital CT scan classification, which directly addresses domain shift and catastrophic forgetting.

\subsection{Vision Transformers and CNNs in Medical Imaging}
Vision transformers, such as the Swin Transformer \cite{liu2021swin}, have recently gained traction in medical imaging due to their superior performance on various tasks and ability to model long-range dependencies. ResNet architectures \cite{he2016deep}, on the other hand, remain popular and reliable baselines. Comparing CL techniques across these architectures provides insight into their adaptability and robustness under domain shift, which is crucial for clinical deployment \cite{hatamizadeh2022unetr}.

\section{Methodology}
\label{sec:method}
Our proposed framework mitigates catastrophic forgetting by developing and applying a novel, patient-aware replay strategy. The core principle is to construct a memory buffer that is not only populated with samples at high risk of being forgotten but that also reflects the hierarchical structure of clinical data, ensuring diversity and relevance. Our methodology is a three-stage process designed for clarity and causal correctness, involving (1) a forgetting analysis stage to quantify representational instability; (2) a patient-aware buffer construction stage using the derived instability signal; and (3) a final drift-guided continual learning stage.

\subsection{Forgetting Analysis and Multi-Layer Drift Quantification} To derive an effective signal for our selection strategy, we first conduct an analytical stage to identify which source-domain samples are most representationally unstable.

\subsubsection{Baseline Model Generation}The process begins with the generation of two essential models:

\textbf{Source Model ($M_A$):} A base model is trained until convergence on the source domain dataset, $D_A$ (Hospital 1). Its feature extractor is denoted by $\phi_A$. This model represents the "ground truth" knowledge we wish to preserve

\textbf{Naively Fine-tuned Model ($M_B$) :} A copy of the source model $M_A$ is then directly fine-tuned on the target domain, $D_B$ (Hospital 2). The resulting model, ($M_B$) with feature extractor $\phi_B$, serves as a "forgetful" model that demonstrates the effects of catastrophic forgetting.

\subsubsection{Multi-Layer Latent Drift Calculation} We define Latent Drift as the change in a model's internal feature representation for a given sample when the model is adapted to a new domain. A large drift signifies that the model's understanding of the sample has been corrupted, marking it as "forgotten." To create a robust measure, we propose Multi-Layer Latent Drift (MLD), which has two key properties.

First, instead of relying on features from a single layer, which can be noisy or overly specific, MLD aggregates information from the final two layers of the network backbone ($L$ and $L-1$). This captures changes at multiple levels of semantic abstraction, providing a more stable and holistic measure of representational change.

Second, we use Cosine Distance as our distance metric. This is a deliberate choice over alternatives like Euclidean distance because it is invariant to the magnitude of the feature vectors. Cosine distance measures the change in the orientation of the vectors, which is a better proxy for a shift in semantic meaning, whereas magnitude can be influenced by unrelated factors like model confidence or calibration.

The MLD for a source-domain sample $x_i$ is formally defined as the average cosine distance across the selected layers \cite{salton1983introduction, manning2008introduction}:


\begin{equation}  \label{MLD-loss} 
\text{MLD}(x_i) = \frac{1}{2} \sum_{l=L-1}^{L} \left( 1 - \frac{\phi_A^l(x_i) \cdot \phi_B^l(x_i)}{\|\phi_A^l(x_i)\|_2 \cdot \|\phi_B^l(x_i)\|_2} \right)
\end{equation}

 where $\phi^l{(x_i)}$ is the feature vector from layer l and L is the final layer index.  This averaging makes the drift score more stable and less sensitive to noise in a single layer. A high MLD score identifies a sample as having high representational instability and thus a high risk of being forgotten.
 
 \subsection{Proposed Buffer Strategy: Patient-Aware Selection}
 This stage is the core of our proposed method. Instead of selecting top-scoring slices globally, which could lead to oversampling from a few patients, our strategy enforces diversity by operating at the patient level. The full procedure is detailed in Algorithm-1

 \begin{itemize}
 
\item \textit{Per-Patient Slice Ranking:} For each patient in the source training set, we rank all of their associated slices based on the Multi-Layer Latent Drift (MLD) scores calculated in forgetting analysis stage.

\item \textit{Buffer Population}: The memory buffer is constructed by selecting a fixed number of the highest-ranked slices (the top 30 slices) from each patient. These sets of slices are then added to the buffer $\mathrm{B}$, starting with patients who exhibit the highest overall average drift, until the desired total buffer size is reached. 
\end{itemize}

This patient-aware approach ensures that the replay buffer contains high-fidelity raw images from a wide array of the most "forgotten" clinical cases, providing a diverse and highly informative dataset for replay.
\\

\hrule
\textbf{Algorithm-1: Patient-Aware Buffer Construction} \label{alg:buffer_construction}
\hrule

\textbf{Input:} Source training set $\mathcal{D}_A^{train}$, MLD scores for all slices, patient IDs, buffer size $K$, slices per patient $S_p = 30$. \\
\textbf{Output:} Memory Buffer $B$.

\begin{algorithmic}[1]
\State Initialize $B \gets \emptyset$
\State Group all slices in $\mathcal{D}_A^{train}$ by patient ID
\For{each patient $P_j$}
    \State Compute average MLD score: $\overline{\text{MLD}}(P_j)$
\EndFor
\State Create a ranked list $P_{ranked}$ of patients sorted by descending $\overline{\text{MLD}}$
\For{each patient $P_j$ in $P_{ranked}$}
    \If{$|B| \geq K$}
        \State \textbf{break}
    \EndIf
    \State Get all slices $\{x_i\}$ belonging to $P_j$
    \State Rank these slices by individual MLD scores (descending)
    \State Select top $S_p$ slices $\{x_j^*\}$
    \State Add the raw images and labels of $\{x_j^*\}$ to $B$
\EndFor
\State \Return $B$
\end{algorithmic}
\hrule

\subsection{Drift-Guided Continual Learning} With the patient-aware intelligent buffer $\mathrm{B}$
 constructed, we perform the final continual learning training run.

 \begin{itemize}
     \item Initialization: We begin again with a fresh instance of the converged source model, $M_A$.

     \item Replay-based Training: The model is trained on the target domain dataset, $D_B$ ,with each mini-batch comprising both new samples from $D_B$ and replayed samples from our buffer $\mathrm{B}$.

    \item Loss Optimization: The model is optimized using a combined loss function that balances learning on the new task and retaining knowledge from the replayed data.

 \end{itemize}

This methodology, centered on patient-aware selection using a multi-layer drift signal, directly targets the mechanisms of forgetting in a way that is tailored to the structure of real-world medical data. The effectiveness of this proposed strategy is compared against simpler baselines and ablations in Section \ref{sec:experiments}.

\section{Dataset}
\label{sec:data}

To evaluate the robustness and fairness of continual learning models in medical imaging, we utilize a curated dataset of chest CT scans collected from two distinct hospitals and medical centers \cite{10678654, KOLLIAS2023126244, 9816321}. Each scan is annotated  as either Covid-19 positive or Normal, which consist the diagnostic labels. 

 
The dataset is partitioned into training, validation, and test subsets. All partitions include scans from these two sources, allowing us to simulate domain shift and assess cross-institution generalization, a key requirement in real-world deployment of continual learning models.

This benchmark is designed to test whether models trained in a continual learning setting can retain diagnostic performance when exposed to data from new or revisited sources. 
To this end, model performance is evaluated using data from Hospitals 1 and 2, where we have a large number of samples. The data from Hospital-1 (H1) is used to train the selected pre-trained backbones creating our source model which consists the baseline one. This model is then fine-tuned on data from Hospital-2 (H2). This approach aims to mitigate the tendency of the model to forget previously learned knowledge.

Our data consist of CT scans that contain multiple slices corresponding to the area of examination. 

\begin{figure}[htbp]
    \includegraphics[width=0.5\textwidth]{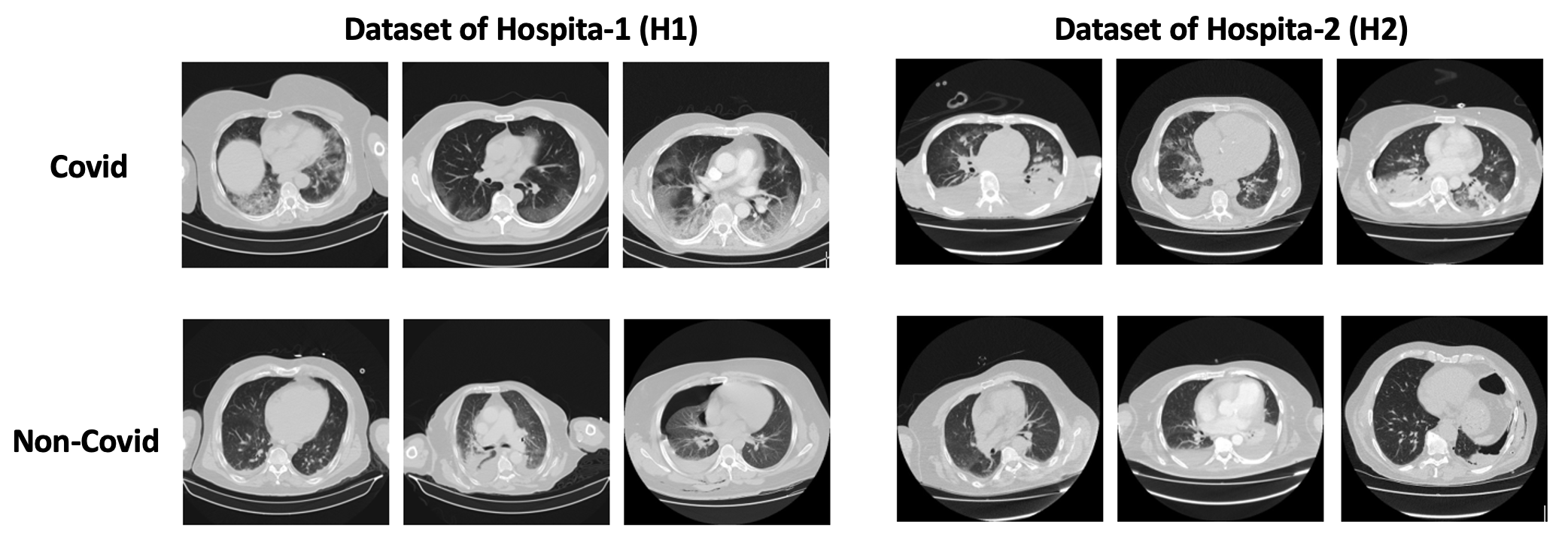} 
    \caption{Samples of CT slices from our datasets.}
    \label{fig:dataset}
\end{figure}

These datasets, H1 and H2, are more appropriate for our task because they provide an adequate number of samples in terms of both patients and slices, thus supplying the necessary information for our models. An indicative sample of these data is found in Figure-\ref{fig:dataset}. The number of the samples related to dataset H1 and H2 are presented on Tables-\ref{tab:h1_data} and \ref{tab:h4_data} respectively. 

Additionally, we observe that both datasets suffer from class imbalance. Hospital-2 (H2) has very few COVID samples compared to non-COVID ones, leading to an imbalance of over 10\%–90\%, while Hospital-1 (H1) overrepresents the COVID class, though with a less severe imbalance of approximately 60\%–40\%.

\begin{table}[]
\centering
\footnotesize  
\setlength{\tabcolsep}{2.5pt}   
\renewcommand{\arraystretch}{1.0}  
\begin{tabular}{|c|ccc|ccc|}
\hline
\multirow{2}{*}{\textbf{H1 Dataset}} & \multicolumn{3}{c|}{\textbf{Patients}}                                                                                   & \multicolumn{3}{c|}{\textbf{Slices}}                                                                                            \\ \cline{2-7} 
                                  & \multicolumn{1}{c|}{\textit{\textbf{Total}}} & \multicolumn{1}{c|}{\textit{\textbf{NonCovid}}} & \textit{\textbf{Covid}} & \multicolumn{1}{c|}{\textit{\textbf{Total}}} & \multicolumn{1}{c|}{\textit{\textbf{NonCovid}}} & \textit{\textbf{Covid}} \\ \hline
\textbf{Training set}             & \multicolumn{1}{c|}{1,230}                    & \multicolumn{1}{c|}{484}                        & 746                     & \multicolumn{1}{c|}{331069}                         & \multicolumn{1}{c|}{132,868}                     & 198,201                  \\ \hline
\textbf{Validation set}           & \multicolumn{1}{c|}{258}                     & \multicolumn{1}{c|}{101}                        & 157                     & \multicolumn{1}{c|}{69,769}                          & \multicolumn{1}{c|}{27,757}                      & 42,012                   \\ \hline
\textbf{Testing set}              & \multicolumn{1}{c|}{281}                     & \multicolumn{1}{c|}{113}                        & 168                     & \multicolumn{1}{c|}{77,826}                          & \multicolumn{1}{c|}{30,990}                      & 46,836                   \\ \hline
\textbf{Total}                    & \multicolumn{1}{c|}{1,769}                    & \multicolumn{1}{c|}{698}                        & 1,071                    & \multicolumn{1}{c|}{478,664}                         & \multicolumn{1}{c|}{191,615}                     & 287,049                  \\ \hline
\end{tabular}
\caption{Dataset of Hospital-1 (H1).}
\label{tab:h1_data}
\end{table}

\begin{table}[]
\centering
\footnotesize  
\setlength{\tabcolsep}{2.5pt}   
\renewcommand{\arraystretch}{1.0}  
\begin{tabular}{|c|ccc|ccc|}
\hline
\multirow{2}{*}{\textbf{H2 Dataset}} & \multicolumn{3}{c|}{\textbf{Patients}}                                                                                   & \multicolumn{3}{c|}{\textbf{Slices}}                                                                                            \\ \cline{2-7} 
                                  & \multicolumn{1}{c|}{\textit{\textbf{Total}}} & \multicolumn{1}{c|}{\textit{\textbf{NonCovid}}} & \textit{\textbf{Covid}} & \multicolumn{1}{c|}{\textit{\textbf{Total slices}}} & \multicolumn{1}{c|}{\textit{\textbf{NonCovid}}} & \textit{\textbf{Covid}} \\ \hline
\textbf{Training set}             & \multicolumn{1}{c|}{1,998}                   & \multicolumn{1}{c|}{1,795}                      & 203                     & \multicolumn{1}{c|}{359,954}                        & \multicolumn{1}{c|}{324,309}                    & 35,645                  \\ \hline
\textbf{Validation set}           & \multicolumn{1}{c|}{420}                     & \multicolumn{1}{c|}{379}                        & 41                      & \multicolumn{1}{c|}{73,814}                         & \multicolumn{1}{c|}{67,033}                     & 6,781                   \\ \hline
\textbf{Testing set}              & \multicolumn{1}{c|}{448}                     & \multicolumn{1}{c|}{395}                        & 53                      & \multicolumn{1}{c|}{79,201}                         & \multicolumn{1}{c|}{69,815}                     & 9,386                   \\ \hline
\textbf{Total}                    & \multicolumn{1}{c|}{2,866}                   & \multicolumn{1}{c|}{2,569}                      & 297                     & \multicolumn{1}{c|}{512,969}                        & \multicolumn{1}{c|}{461,157}                    & 51,812                  \\ \hline
\end{tabular}
\caption{Dataset of Hospital-2 (H2).}
\label{tab:h4_data}

\end{table}

\section{Experiments}
\label{sec:experiments}
To rigorously evaluate our proposed continual learning framework, we designed a series of experiments to dissect the impact of different replay strategies on mitigating catastrophic forgetting under domain shift. Our evaluation is structured to answer three primary research questions 1. How does our proposed latent drift-guided replay compare against standard CL baselines? 2. What are the specific contributions of patient-aware selection and multi-layer drift aggregation to performance? 3. How do modern Transformer and traditional CNN architectures respond to these strategies?

\subsection{Experimental Setup}
\label{sec:strategies}

\subsubsection{Datasets and Continual Learning Task} We use a curated dataset of chest CT scans for COVID-19 classification, aggregated from multiple real-world hospitals and medical centers \cite{10678654, KOLLIAS2023126244, 9816321}. Our continual learning scenario simulates a practical domain shift, where a model is first trained on the source domain, Hospital-1 (H1), and then must adapt to the target domain, Hospital-2 (H2). These two domains were chosen for their significant size and pronounced differences in class distribution (H1: $\sim$60\% COVID; H2: $\sim$10\% COVID) and imaging characteristics, providing a challenging and realistic testbed. All datasets were split into training, validation, and testing sets at the patient level to prevent data leakage.

\subsubsection{Models and Implementation Details}
\textbf{Architectures:} We use two powerful, widely-adopted backbones pre-trained on ImageNet: ResNet50 \cite{he2016deep}, representing convolutional neural networks (CNNs), and Swin Transformer \cite{liu2021swin}, representing state-of-the-art Vision Transformers.

\textbf{Training Protocol:} All models were trained using the AdamW optimizer, with a learning rate of $5e-5$ for Swin Transformer and $1e-5$ for ResNet50.  The batch size was set to 32. The initial training on the H1 source domain was conducted for 15 epochs. The subsequent continual learning phase on the H2 target domain was run for 10 epochs.

\textbf{Handling Data Imbalance:} To address the severe class imbalance, we employed a combination of a Weighted Random Sampler \cite{Efraimidis2008} at the data loader level and a Focal Loss \cite{lin2017focal} function during training.

\textbf{Data Augmentation:} Standard data augmentation techniques, including random horizontal flips and rotations, were applied during training to improve model generalization.

\textbf{Replay Buffer Configuration:} For all replay-based experiments, the memory buffer $\mathrm{B}$ was configured with a fixed capacity of 30,000 samples, corresponding to approximately 10\% of the H1 training set. To prevent bias from the imbalanced source data, the buffer was explicitly class-balanced with 15,000 samples from the COVID class and 15,000 from the non-COVID class. During replay, mini-batches were constructed with a 50\% probability of drawing from the H2 training set or the H1 buffer.

\subsection{Continual Learning Strategies}
We systematically evaluate a comprehensive set of strategies, organized to allow for direct comparison and ablation. Each strategy defines a method for constructing the replay buffer, resulting in a distinct final model.

\subsubsection{Group A: Baseline Strategies --
These models serve as fundamental points of comparison}

\textbf{Source-Only:} The backbone is finetuned only on H1. This model establishes the upper bound for source domain performance and quantifies the initial domain gap when tested on H2.

\textbf{Naive Fine-tuning:} The source model is further fine-tuned on H2 without any replay mechanism. This serves as the lower bound for retention, demonstrating catastrophic forgetting.

\textbf{Random Replay:} The buffer is populated by randomly sampling 30,000 class-balanced samples from the H1 training set. This is the standard and most common replay baseline.

\subsubsection{Group B: Latent Drift-Guided Strategies}
These models leverage the Latent Drift (LD) signal, as defined in Section \ref{sec:method}, to inform buffer selection. We explore variations to test key hypotheses.

\textbf{--Our Proposed Method--}

\textbf{Patient-Aware Multi-Layer Drift:} This is our main proposed strategy. The buffer is populated by selecting the 30 slices with the highest Multi-Layer Latent Drift (MLD) from each of the top-ranked patients, as detailed in Section- \ref{sec:method}.

\textbf{--Ablation Studies--}

\textbf{Patient-Aware vs. Alternative Selection Criteria: }

Global Slice and Center Slice – To isolate the benefit of the patient-aware approach, we compare our proposed model against global and center-slice MLD selection. In the Global Slice variant, the buffer stores the 30,000 slices with the highest MLD scores, selected from the entire H1 training set irrespective of patient origin. In the Center Slice variant, to test whether focusing on anatomically central slices is beneficial, we evaluate versions of our core strategy that restrict selection to central slices only.

\textbf{Multi-Layer vs. Single-Layer Drift: } To validate the use of a multi-layer signal, we compare our proposed Model against Patient-Aware Single-Layer Drift: A variant of our proposed method that uses LD calculated from only the final backbone layer.

\textbf{Choice of Distance Metric:} To analyze the sensitivity to the drift metric, we replace the Cosine Distance in our MLD calculation with L2 Euclidean Distance and Mahalanobis Distance.

\textbf{Hybrid Drift and Uncertainty:} To investigate the synergy between drift and model uncertainty, we test Drift \& Entropy: The buffer is populated based on a combined score.

The entropy is referred to the softmax output of the current model used for fine-tuning for each slice.

Given the softmax probability vector \(\mathbf{p} = (p_1, p_2, \ldots, p_C)\) over \(C\) classes, the entropy \(H(\mathbf{p})\) is defined as:
\[
H(\mathbf{p}) = - \sum_{i=1}^C p_i \log p_i
\]
where \(p_i\) is the predicted probability for class \(i\). The entropy measures the uncertainty of the model's prediction, with higher values indicating greater uncertainty.

To select slices based on both uncertainty and latent drift score \(D\), a combined score \(S\) can be computed as:
\[
S = \alpha \cdot D + \beta \cdot H(\mathbf{p})
\]
where \(\alpha, \beta \geq 0\) are weighting factors balancing the contribution of uncertainty and drift. Slices with higher values of \(S\) are prioritized for further analysis or labeling. In our case, we set \(\alpha = 0.7  \) and \(\beta = 0.3 \) as determined empirically through a validation set.

\subsection{Evaluation Metrics}
The performance of each strategy is evaluated from multiple perspectives:

\textbf{Task Performance (Accuracy):} This is our primary measure of model effectiveness. We report classification accuracy on the held-out test sets of H1 (to measure knowledge retention) and H2 (to measure adaptation). Performance is reported both per-slice and per-patient (via majority vote) to reflect both granular and clinical-level diagnostic accuracy.

\textbf{Forgetting and Transfer:} We use standard CL metrics \cite{kumari2024continuallearningmedicalimage}, including Backward Transfer (BWT) and Forward Transfer (FWT). 

BWT measures the performance change on the source task after learning the target task. A BWT score closer to zero indicates less forgetting.
\[
\text{BWT}_i = R_{j,i} - R_{i,i}, \quad \text{where} \quad j > i
\]

$R_{i,j}$: Accuracy on task $j$ after training up to task $i$, $R^0_j$: Initial accuracy on task $j$ before any CL training, $R_{i,i}$: Accuracy on task $i$ immediately after training task $i$ and $R_{j,i}$: Accuracy on task $i$ after training up to task $j$.

 FWT measures how learning previous tasks improves performance on a new task before it has been trained. Positive FWT indicates that knowledge learned earlier helps with future tasks.

\[
\text{FWT}_j = R_{i,j} - R^0_j, \quad \text{where} \quad i < j
\]

\textbf{Representational Stability:} We use the Latent Representation Shift (LRS), defined as the average MLD on the H1 test set between the source model and the final continual learning model. A lower LRS score signifies superior preservation of the original feature representations.

All experiments were run on a server with 8 × NVIDIA Tesla V100 GPUs. 

\section{Results}
\label{sec:results}

\begin{table*}[h!]
\centering
\resizebox{\textwidth}{!}{%
\begin{tabular}{|l|cccc|cccc|}
\hline
 & \multicolumn{4}{c|}{\textbf{Accuracy Per-Patient (\%)}} & \multicolumn{4}{c|}{\textbf{Accuracy Per-Slice (\%)}}  \\ \hline
\textbf{Model} & \multicolumn{1}{c|}{\textbf{SwinT-H2}} & \multicolumn{1}{c|}{\textbf{SwinT-H1}} & \multicolumn{1}{c|}{\textbf{ResNet-H2}} & \textbf{ResNet-H1} & \multicolumn{1}{c|}{\textbf{SwinT-H2}} & \multicolumn{1}{c|}{\textbf{SwinT-H1}} & \multicolumn{1}{c|}{\textbf{ResNet-H2}} & \textbf{ResNet-H1} \\ \hline
\textbf{Source-Only   (No CL)} & \multicolumn{1}{c|}{88.17} & \multicolumn{1}{c|}{94.24} & \multicolumn{1}{c|}{88.66} & 93.17 & \multicolumn{1}{c|}{81.60} & \multicolumn{1}{c|}{93.27} & \multicolumn{1}{c|}{82.55} & 91.94  \\ \hline
\textbf{Naive Fine-tuning} & \multicolumn{1}{c|}{95.31} & \multicolumn{1}{c|}{51.80} & \multicolumn{1}{c|}{69.87} & 71.58 & \multicolumn{1}{c|}{93.62} & \multicolumn{1}{c|}{58.79} & \multicolumn{1}{c|}{90.41} & 66.7  \\ \hline
\textbf{Random Replay} & \multicolumn{1}{c|}{90.70} & \multicolumn{1}{c|}{92.21} & \multicolumn{1}{c|}{79.01} & 87.29 & \multicolumn{1}{c|}{83.35} & \multicolumn{1}{c|}{87.82} & \multicolumn{1}{c|}{78.86} & 87.53  \\ \hline
\textbf{Proposed Method} 
& \multicolumn{1}{c|}{\textbf{93.75}} & \multicolumn{1}{c|}{\textbf{92.45}} & \multicolumn{1}{c|}{\textbf{89.29}} & \textbf{88.13} & \multicolumn{1}{c|}{\textbf{90.19}} & \multicolumn{1}{c|}{\textbf{89.54}} & \multicolumn{1}{c|}{\textbf{81.52}} & \textbf{87.00}  \\ \hline
\end{tabular}%
}
\caption{Comparison of our proposed method against key baselines. The results are presented for both per-patient and per-slice accuracy on the target (H2) and source (H1) hospitals.}
\label{tab:main-proposed}
\end{table*}

\begin{table*}[]
\resizebox{\textwidth}{!}{%
\begin{tabular}{|l|cccc|cccc|}
\hline

{\textbf{\begin{tabular}[c]{@{}l@{}}Patient Aware \\ vs Alternative Selection\end{tabular}}} 
& \multicolumn{4}{c|}{\textbf{Accuracy   Per-Patient (\%)}} & \multicolumn{4}{c|}{\textbf{Accuracy   Per-Slice (\%)}}  \\ \hline

\textbf{Model} & \multicolumn{1}{c|}{\textbf{SwinT-H2}} & \multicolumn{1}{c|}{\textbf{SwinT-H1}} & \multicolumn{1}{c|}{\textbf{ResNet-H2}} & \textbf{ResNet-H1} & \multicolumn{1}{c|}{\textbf{SwinT-H2}} & \multicolumn{1}{c|}{\textbf{SwinT-H1}} & \multicolumn{1}{c|}{\textbf{ResNet-H2}} & \textbf{ResNet-H1} \\ \hline
{\textbf{Global Slice Selection}} & \multicolumn{1}{c|}{91.74} & \multicolumn{1}{c|}{82.01} & \multicolumn{1}{c|}{84.82} & 87.05 & \multicolumn{1}{c|}{87.53} & \multicolumn{1}{c|}{80.2} & \multicolumn{1}{c|}{80.75} & 86.21  \\ \hline
{\textbf{Center Slice Selection}} & \multicolumn{1}{c|}{93.97} & \multicolumn{1}{c|}{92.09} & \multicolumn{1}{c|}{81.03} & 87.41 & \multicolumn{1}{c|}{82.34} & \multicolumn{1}{c|}{72.49} & \multicolumn{1}{c|}{85.99} & 81.95  \\ \hline
{\textbf{Proposed Method}} & \multicolumn{1}{c|}{\textbf{93.75}} & \multicolumn{1}{c|}{\textbf{92.45}} & \multicolumn{1}{c|}{\textbf{89.29}} & \textbf{88.13} & \multicolumn{1}{c|}{\textbf{90.19}} & \multicolumn{1}{c|}{\textbf{89.54}} & \multicolumn{1}{c|}{\textbf{81.52}} & \textbf{87.00}  \\ \hline
\end{tabular}%
}
\caption{Impact of Patient-Aware vs. Alternative Selection, Global and Center. The results are presented for both per-patient and per-slice accuracy on the target (H2) and source (H1) hospitals.}
\label{tab:Impact of PatientAware-Global Selection}
\end{table*}

\begin{table*}[h!]
\centering
\resizebox{\textwidth}{!}{%
\begin{tabular}{|l|cccc|cccc|}
\hline
{\textbf{\begin{tabular}[c]{@{}l@{}}Multi-Layer \\ vs Single-Layer Drift\end{tabular}}} 

& \multicolumn{4}{c|}{\textbf{Accuracy   Per-Patient (\%)}} & \multicolumn{4}{c|}{\textbf{Accuracy   Per-Slice (\%)}}  \\ \hline
\textbf{Model} & \multicolumn{1}{c|}{\textbf{SwinT-H2}} & \multicolumn{1}{c|}{\textbf{SwinT-H1}} & \multicolumn{1}{c|}{\textbf{ResNet-H2}} & \textbf{ResNet-H1} & \multicolumn{1}{c|}{\textbf{SwinT-H2}} & \multicolumn{1}{c|}{\textbf{SwinT-H1}} & \multicolumn{1}{c|}{\textbf{ResNet-H2}} & \textbf{ResNet-H1} \\ \hline
{\textbf{Single-Layer   Drift}}  & \multicolumn{1}{c|}{93.51} & \multicolumn{1}{c|}{91.37} & \multicolumn{1}{c|}{81.03} & 90.29 & \multicolumn{1}{c|}{84.14} & \multicolumn{1}{c|}{87.67} & \multicolumn{1}{c|}{76.98} & 86.19  \\ \hline
{\textbf{Proposed Method}} & \multicolumn{1}{c|}{\textbf{93.75}} & \multicolumn{1}{c|}{\textbf{92.45}} & \multicolumn{1}{c|}{\textbf{89.29}} & \textbf{88.13} & \multicolumn{1}{c|}{\textbf{90.19}} & \multicolumn{1}{c|}{\textbf{89.54}} & \multicolumn{1}{c|}{\textbf{81.52}} & \textbf{87.00}  \\ \hline
\end{tabular}%
}
\caption{Impact of proposed Multi-Layer vs. Single-Layer Drift. The results are presented for both per-patient and per-slice accuracy on the target (H2) and source (H1) hospitals.}
\label{tab:Impact of Multi-Single Layer Drift}
\end{table*}

We present a comprehensive analysis of our experimental results. The findings are organized to first establish the baseline performance, then to dissect the specific contributions of our proposed methodology through targeted ablations, and finally to discuss broader architectural implications. All results are presented at both the per-slice level, to assess granular feature learning, and the per-patient level, which reflects a more realistic clinical diagnostic workflow.

\subsection{Main Finding: Drift-Guided Replay Prevents Catastrophic Forgetting}
The results presented in Table-\ref{tab:main-proposed} shows that our proposed method achieves the strongest stability–plasticity trade-off across backbones. On Swin Transformer, our proposed method attains 92.45\% (H1) / 93.75\% (H2)—improving over Random Replay by +0.24 percentage points (pp) on H1 and +3.05 pp on H2, while remaining within 1.8 pp of Source-Only on H1 but substantially higher on H2. On ResNet-50, our proposed method reaches 88.13\% / 89.29\%, yielding the highest H2 overall and surpassing Random Replay by +0.84 pp (H1) and +10.28 pp (H2). As a simple balance metric, our proposed method also maximizes min(H1,H2) among continual-learning (CL) strategies for both backbones (Swin: 92.45; ResNet-50: 88.13), indicating robust retention without sacrificing adaptation. Per-slice results mirror per-patient trends and are included for completeness. On Swin, our proposed method improves over Random Replay by +1.72 pp (H1) / +6.84 pp (H2); on ResNet-50, a small -0.53 pp on H1 is offset by +2.66 pp on H2, maintaining the same overall ranking.

\begin{figure}[t]
  \centering
  \includegraphics[width=\columnwidth]{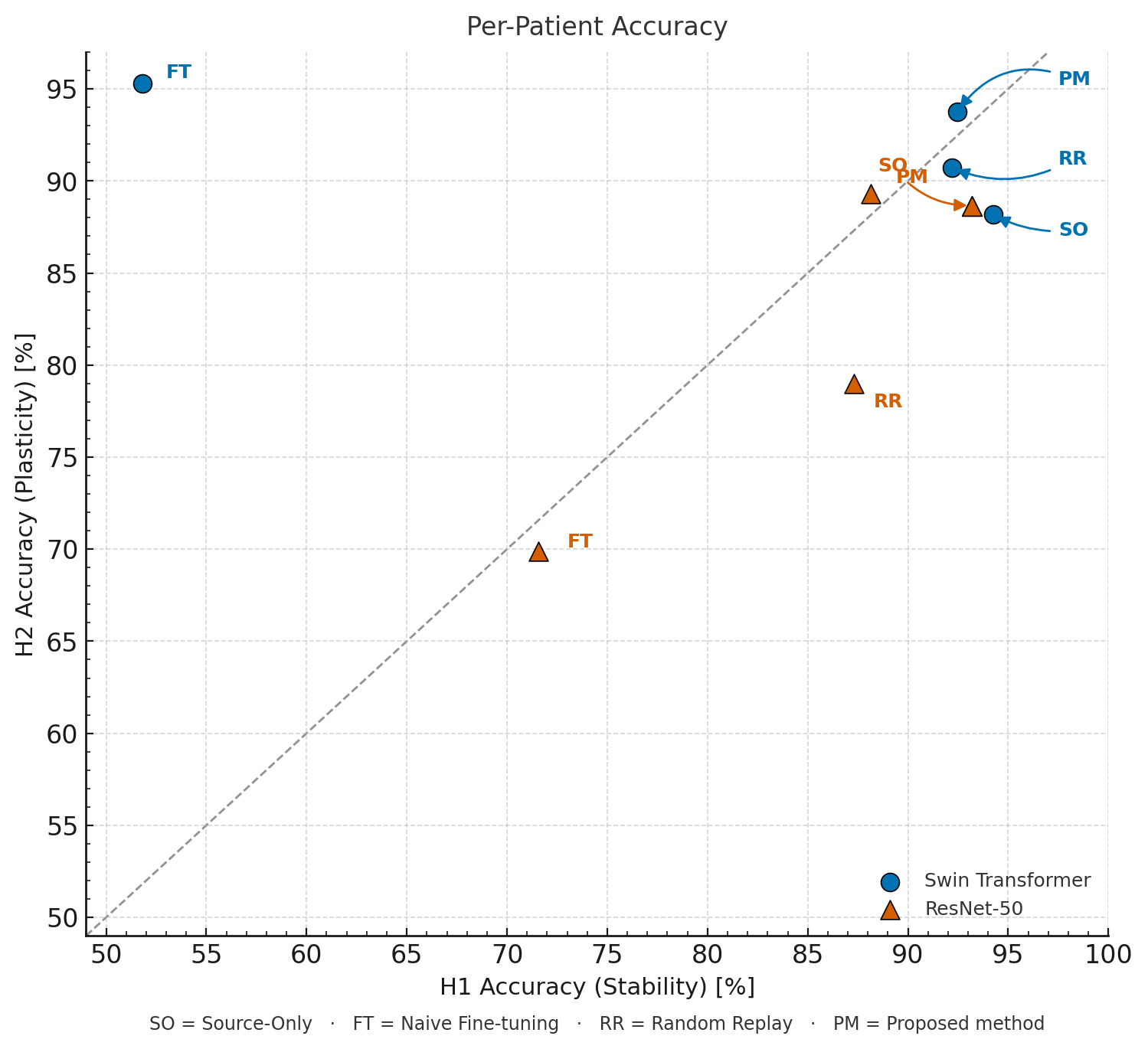}
  \vspace{-4pt}
  \caption{Stability–plasticity trade-off (H1 vs H2), dashed line denotes equal stability/plasticity.}
  \label{fig:stability-plasticity-per-patient}
  \vspace{-6pt}
\end{figure}

Figure-\ref{fig:stability-plasticity-per-patient}, the per-patient stability–plasticity scatter confirms this trade-off: the dashed diagonal denotes equal stability and plasticity. Naive fine-tuning lies above the line (plastic but forgetful), Source-Only below (stable but under-adaptive), Random Replay moves toward the line, and our proposed method sits closest to the top-right region for both backbones, visually reflecting the best joint performance. 

\begin{table*}[h!]
\resizebox{\textwidth}{!}{%
\centering
\begin{tabular}{|l|cccc|cccc|}
\hline
{\textbf{Alternative   Strategies}} & \multicolumn{4}{c|}{\textbf{Accuracy   Per-Patient (\%)}} & \multicolumn{4}{c|}{\textbf{Accuracy   Per-Slice (\%)}}  \\ \hline
\textbf{Model} & \multicolumn{1}{c|}{\textbf{SwinT-H2}} & \multicolumn{1}{c|}{\textbf{SwinT-H1}} & \multicolumn{1}{c|}{\textbf{ResNet-H2}} & \textbf{ResNet-H1} & \multicolumn{1}{c|}{\textbf{SwinT-H2}} & \multicolumn{1}{c|}{\textbf{SwinT-H1}} & \multicolumn{1}{c|}{\textbf{ResNet-H2}} & \textbf{ResNet-H1} \\ \hline
\textbf{Euclidean Distance} & \multicolumn{1}{c|}{94.42} & \multicolumn{1}{c|}{90.29} & \multicolumn{1}{c|}{85.71} & 89.93 & \multicolumn{1}{c|}{88.40} & \multicolumn{1}{c|}{88.06} & \multicolumn{1}{c|}{71.91} & 85.81  \\ \hline
\textbf{Mahalanobis Distance} & \multicolumn{1}{c|}{91.74} & \multicolumn{1}{c|}{91.01} & \multicolumn{1}{c|}{77.23} & 86.69 & \multicolumn{1}{c|}{80.32} & \multicolumn{1}{c|}{87.78} & \multicolumn{1}{c|}{75.93} & 84.92  \\ \hline
{\color[HTML]{1A1C1E} \textbf{Latent drift  \& Entropy}} & \multicolumn{1}{c|}{92.86} & \multicolumn{1}{c|}{91.07} & \multicolumn{1}{c|}{91.01} & 84.17 & \multicolumn{1}{c|}{79.97} & \multicolumn{1}{c|}{83.48} & \multicolumn{1}{c|}{79.79} & 82.99  \\ \hline
\textbf{Proposed Method} & \multicolumn{1}{c|}{\textbf{93.75}} & \multicolumn{1}{c|}{\textbf{92.45}} & \multicolumn{1}{c|}{\textbf{89.29}} & \textbf{88.13} & \multicolumn{1}{c|}{\textbf{90.19}} & \multicolumn{1}{c|}{\textbf{89.54}} & \multicolumn{1}{c|}{\textbf{81.52}} & \textbf{87.00}  \\ \hline
\end{tabular}%
}
\caption{Performance of alternative strategies. The results are presented for both per-patient and per-slice accuracy on the target (H2) and source (H1) hospitals.}
\label{tab:Alternative Strategies}
\end{table*}

\subsection{Dissecting the Methodology: Analysis of Design Choices}
Our ablation studies validate the key architectural decisions behind our proposed method.

\subsubsection{The Superiority of Patient-Aware Selection} As shown in Table-\ref{tab:Impact of PatientAware-Global Selection}, the choice between a global or center slice-level selection and our patient-aware approach has a profound impact on knowledge retention. While the Center and Global Slice Selection methods improve over random replay, our Proposed Method is substantially better at preserving H1 performance. For the Swin Transformer, patient-aware selection boosts per-patient H1 accuracy from 82.01\% of global and 92.09\% for center to a remarkable 92.45\%.  This confirms our hypothesis that a global strategy risks creating a redundant buffer by over-sampling from a few "difficult" patients. Center-slice selection seems to give better results, given that these slices may be more informative in a CT-scan, but finally doesn't achieve overall the highest performance. In contrast, our patient-aware approach ensures a more diverse and efficient memory by sampling from a wider range of high-drift clinical cases.

\subsubsection{The Benefit of a Multi-Layer Drift Signal}
The comparison in Table-\ref{tab:Impact of Multi-Single Layer Drift} demonstrates the value of using a more robust drift signal. The Multi-Layer Drift approach consistently outperforms the Single-Layer Drift strategy in preserving H1 knowledge across both architectures. For example, with the Swin model, using a multi-layer signal improves per-patient H1 retention from 91.37\% to 92.45\%. This suggests that forgetting is a complex process affecting features at multiple levels of semantic abstraction, and a more holistic drift signal is better at identifying samples whose core representations have become unstable.

\subsubsection{Performance of Alternative Strategies}
Our investigation into other advanced strategies, presented in Table-\ref{tab:Alternative Strategies}, provides further context. Using Euclidean Distance proves to be a very strong alternative to Cosine Similarity, achieving the highest H2 accuracy for Swin (94.42\%). This indicates that the core concept of drift-based selection is robust and not overly sensitive to the specific distance metric. The Latent Drift \& Entropy model also performs well, particularly for the ResNet backbone, suggesting that adding an explicit uncertainty signal can benefit less robust architectures. However, for the powerful Swin Transformer, our proposed Cosine-based, pure-drift method delivered the best overall balance, especially in retaining critical knowledge from the source domain.

\begin{table}[!t]
\centering
\resizebox{\columnwidth}{!}{%
\begin{tabular}{|l|c|c||c|c|}
\hline
\multicolumn{1}{|c|}{} & \multicolumn{2}{c|}{\textbf{FWT}} & \multicolumn{2}{c|}{\textbf{BWT (closer to 0)}} \\ \hline
\textbf{Model} & \textbf{SwinT} & \textbf{ResNet-50} & \textbf{SwinT} & \textbf{ResNet-50} \\ \hline
\textbf{Naive Fine-tuning} & \textbf{0.029} & \textbf{0.768} & -0.424 & -0.216 \\ \hline
\textbf{Random Replay}     & \textbf{0.029} & \textbf{0.768} & -0.020 & -0.059 \\ \hline
\textbf{Proposed Method}   & \textbf{0.029} & \textbf{0.768} & \textbf{-0.018} & \textbf{-0.050} \\ \hline
\end{tabular}%
}
\caption{Per-patient Forward Transfer (FWT) and Backward Transfer (BWT).}
\label{tab:FWT-BWT}
\end{table}

\subsection{Analysis of Forgetting and Knowledge Transfer}
To further quantify the learning dynamics, we analyzed the standard continual learning metrics of Forward Transfer (FWT) and Backward Transfer (BWT), with the per-patient results presented in Table-\ref{tab:FWT-BWT}. As expected given the shared H1 initialization, FWT is identical across methods (SwinT: 0.029; ResNet-50: 0.768), confirming that pretraining on H1 provides the same starting benefit for H2. The differentiator is BWT, where higher values (closer to 0) are better: Naive fine-tuning shows severe forgetting (-0.424 on SwinT; -0.216 on ResNet-50). Random Replay substantially reduces forgetting (-0.020 on SwinT; -0.059 on ResNet-50). Our proposed method is best on both backbones, with BWT of -0.018 on SwinT and -0.050 on ResNet-50; this corresponds to improvements over Naive FT of +0.406 and +0.166, and small gains over Random Replay of +0.002 and +0.009, respectively. These transfer metrics align with Table-\ref{tab:main-proposed} and the stability-plasticity plot in Figure-\ref{fig:stability-plasticity-per-patient} preserving H1 performance while adapting to H2, achieving the most favorable balance. Overall, the BWT analysis confirms that our latent drift-guided approach is highly effective at mitigating catastrophic forgetting. 

Beyond the aggregate metrics, it is notable that the absolute magnitude of forgetting achieved by our drift-guided replay is substantially lower than that reported in prior continual learning studies for medical imaging. This demonstrates that a targeted buffer composition can deliver high retention efficiency: with only 10\% of the original source data stored, the proposed method preserves nearly the same H1 performance as models trained with full access to historical data. Such efficiency is particularly relevant in clinical scenarios, where storage, transfer, and privacy constraints limit the feasibility of large replay buffers. Moreover, the latent drift signal itself could serve as a diagnostic indicator, highlighting cases most prone to representational degradation and enabling proactive mitigation before domain adaptation.

\subsection{Architectural Insights}  \textbf{Swin Transformer vs. ResNet50:}
A consistent trend across all tables is the superior performance of the Swin Transformer over ResNet50 in this continual learning scenario. The Swin model not only achieves higher absolute accuracies but also demonstrates greater resilience to forgetting. For example, under naive fine-tuning Table-\ref{tab:main-proposed}, the ResNet model's H1 performance drops less dramatically than Swin's, but its overall performance with our proposed method is significantly lower. We attribute Swin's strength to its hierarchical self-attention mechanism, which is better suited to modeling the long-range, contextual features that define an imaging domain. This makes it more adaptable and, when paired with our intelligent replay strategy, more capable of preserving complex, learned knowledge.

\section{Conclusion and Future work}
\label{sec:conclusion}
In this paper, we addressed the critical challenge of catastrophic forgetting in domain-shifted medical imaging tasks. We demonstrated that naive fine-tuning leads to a severe degradation of prior knowledge, while standard random replay offers only a limited solution. To overcome this, we introduced a novel continual learning framework driven by a patient-aware, latent drift-guided replay strategy.

Our methodology successfully identifies and replays the samples most at risk of being forgotten by quantifying representational instability across multiple feature layers. Through extensive experiments on a real-world, cross-hospital COVID-19 CT dataset, we have shown that our proposed approach significantly outperforms standard baselines. The results validate our core hypotheses: a patient-aware selection strategy is superior to global and center slice-level sampling, a multi-layer drift signal is more robust than a single-layer one, and transformer-based architectures like the Swin Transformer are inherently more resilient to domain shifts than their CNN counterparts.

Overall, this work establishes latent drift as a practical and interpretable signal for constructing highly effective replay memories. Our framework offers a robust and scalable solution that balances stability and plasticity, paving the way for the deployment of truly adaptive AI systems in dynamic clinical environments where trust and reliability are paramount.

\textbf{Future Work:} Our findings open several promising avenues for future research. We plan to extend this methodology to more complex, multi-domain scenarios involving sequences of several different hospitals. Furthermore, we will explore the application of our drift-based selection strategy to 3D volumetric CT data, which may reveal different data structures and forgetting dynamics. Finally, integrating our drift signal with formal uncertainty quantification and exploring its utility in a federated learning setting are exciting directions for developing privacy-preserving, continuously adapting medical AI.

{
    \small
    \bibliographystyle{ieeenat_fullname}
    \bibliography{main}
}

\end{document}